\documentclass[letterpaper, 10 pt, journal, twoside]{IEEEtran}

\IEEEoverridecommandlockouts

\usepackage{amsmath,amssymb,amsfonts}
\usepackage[size=footnotesize]{caption}
\usepackage{algorithmic}
\usepackage{graphicx}
\usepackage{textcomp}
\usepackage{xcolor}

\usepackage{subfig}
\usepackage{dsfont}
\usepackage{perl_acronyms}
\usepackage{soul}
\usepackage{booktabs}
\usepackage[
    style=ieee,
    doi=false,
    isbn=false,
    url=false,
    eprint=false,
    backend=bibtex,
    natbib=true
    ]{biblatex}
\newcommand{\myyu}[1]{}
\renewcommand{\myyu}[1]{{\color{magenta} M-Y: {#1}}}
\newcommand{\matt}[1]{}
\renewcommand{\matt}[1]{{\color{red} MattJR: {#1}}}
\newcommand{\ram}[1]{}
\renewcommand{\ram}[1]{{\color{blue} RAM: {#1}}}
\newcommand{\mattp}[1]{}
\renewcommand{\mattp}[1]{{\color{brown} MattP: {#1}}}
\newcommand{\XD}[1]{}
\renewcommand{\XD}[1]{{\color{orange} XD: {#1}}}

\begin{document}

\title{Risk Assessment and Planning with Bidirectional Reachability for Autonomous Driving}

\author{
    Ming-Yuan Yu$^{1}$, Ram Vasudevan$^{2}$, and Matthew Johnson-Roberson$^{3}$
    \thanks{This work was supported by a grant from Ford Motor Company via the Ford-UM Alliance under Award N022884. (Corresponding author: Ming-Yuan Yu.)}
    \thanks{$^{1}$M.-Y Yu is with Robotics Institute, University of Michigan, Ann Arbor, MI 48109, USA
        {\tt\footnotesize myyu@umich.edu}}%
    \thanks{$^{2} $R. Vasudevan is with the Department of Mechanical Engineering, the University of Michigan, Ann Arbor, MI 48109, USA
        {\tt\footnotesize ramv@umich.edu}}%
    \thanks{$^{3} $M. Johnson-Roberson is with the Department of Naval Architecture and Marine Engineering at the University of Michigan, Ann Arbor, MI 48109, USA
        {\tt\footnotesize mattjr@umich.edu}}%
}

\maketitle

\begin{abstract}
Risk assessment to quantify the danger associated with taking a certain action is critical to navigating safely through crowded urban environments during autonomous driving.
Risk assessment and subsequent planning is usually done by first tracking and predicting trajectories of other agents, such as vehicles and pedestrians, and then choosing an action to avoid future collisions.
However, few existing risk assessment algorithms handle occlusion and other sensory limitations effectively.
One either assesses the risk in the worst-case scenario and thus makes the ego vehicle overly conservative, or predicts as many hidden agents as possible and thus makes the computation intensive.
This paper explores the possibility of efficient risk assessment under occlusion via both forward and backward reachability.
The proposed algorithm can not only identify the location of risk-inducing factors, but can also be used during motion planning.
The proposed method is evaluated on various four-way highly occluded intersections with up to five other vehicles in the scene.
Compared with other risk assessment algorithms, the proposed method shows better efficiency, meaning that the ego vehicle reaches the goal at a higher speed.
In addition, it also lowers the median collision rate by 7.5$\times$ when compared to state of the art techniques.
\end{abstract}

\begin{IEEEkeywords}
Autonomous Vehicle Navigation, Motion and Path Planning, Collision Avoidance
\end{IEEEkeywords}

\section{Introduction}
\label{sec:introduction}

\begin{figure}[t]
  \centering
  \includegraphics[width=0.8\linewidth,trim={5mm 5mm 160mm 13mm},clip]{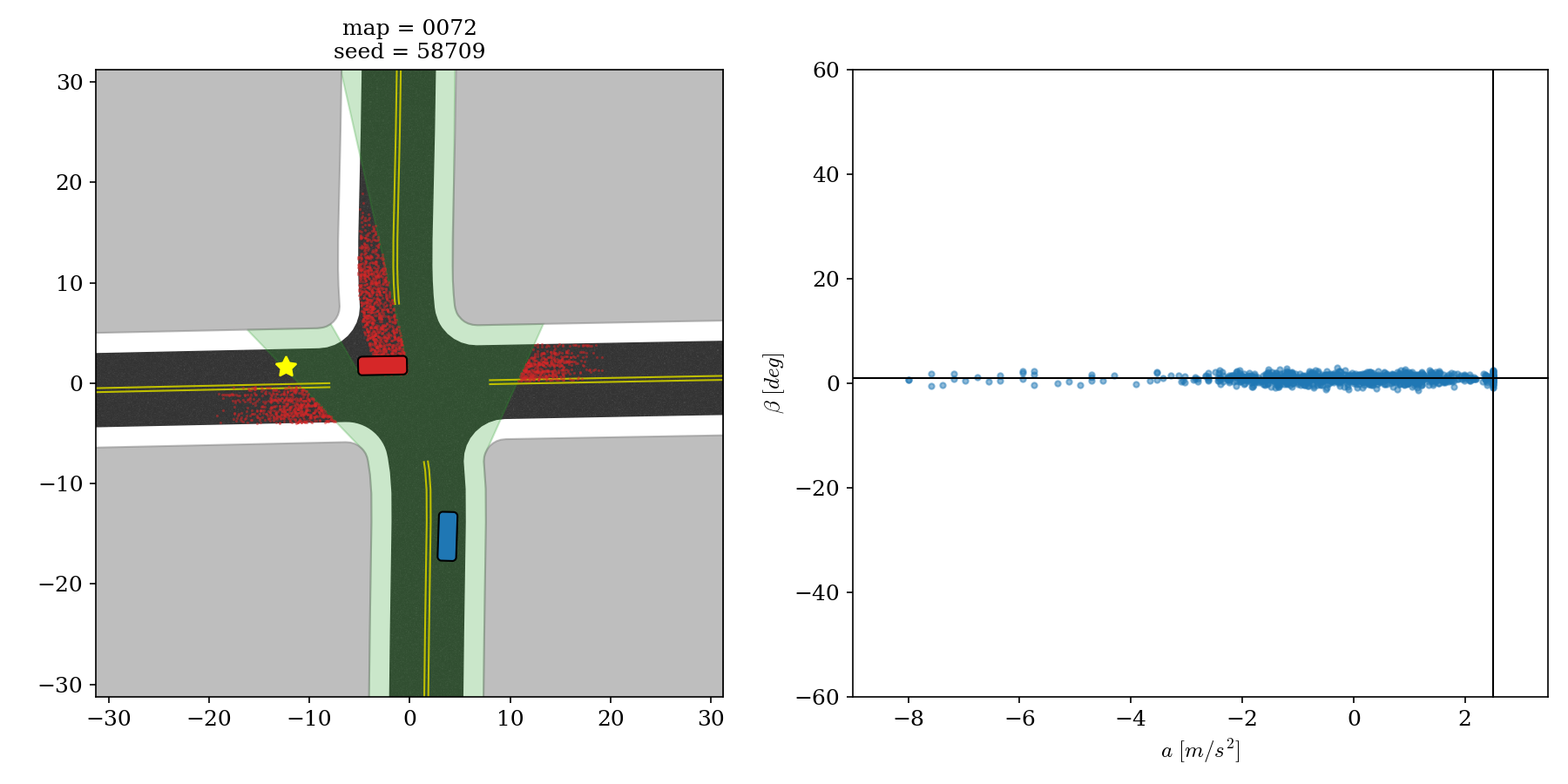}
  \caption{
  The ego vehicle (blue box) intends to perform an unprotected left turn to the goal (yellow star) at an intersection. The green shaded region is the observable polygon and arises due to the 1) limited sensor range and 2) occlusions from another vehicle (red box) and buildings (gray regions.) The proposed algorithm quantifies and identifies the location of risk posed by other vehicles (red particles). Our approach identifies and focuses its attention on risk-inducing regions in the near future (e.g. 1.5 seconds) to ensure that it does not waste computational resources predicting everything in the occluded region. Both axes are in meters.}
  \label{fig:demo_init}
\end{figure}

Most recent autonomous vehicles are equipped with a full sensor suit, including radars, cameras, and \acp{LIDAR}~\cite{kato2015open}. These sensors give the vehicle the ability to perceive the world and to assess the risk of taking a certain action. The quantified risk can then be passed to a planning algorithm to eventually take a action. However, it is still challenging for an autonomous vehicle to navigate through complex environments such as occluded urban intersections such as the one shown in Figure \ref{fig:demo_init}.

The primary challenge of safely navigation through these occluded scenarios is predicting the behavior of all possible incoming traffic.
The challenge arises due to the size of the state space of all possible incoming traffic from every occluded region which can make computation intractable.

Traditionally only forward reachability is used for risk assessment. In other words, algorithms make predictions of all other agents' future trajectories, and select an optimal action based on collision probability \cite{lee2017collision} and time-to-collision \cite{ammoun2009real}.

In contrast, human drivers can usually handle scenarios like the one in Figure \ref{fig:demo_init} even with relatively limited sensing capabilities, such as a narrow field-of-view. We argue that this is because human drivers sense the world in a more efficient way: using semantic and prior information to focus sensing resources on portions of the scene that are most pertinent. For instance, when driving normally, drivers focus mostly on the current and future location of the ego vehicle. We only glimpse to check the rear view mirror when driving forward, but never focus our attention on it for an extended period of time. This is due to the fact that we have 1) a semantic understanding of the environment (e.g. road layout, speed limit, etc) and 2) prior knowledge such as the ability to estimate the reachability of not only for the ego vehicle but also other agents. As a result, we know to focus our limited sensing resources on only the regions where the ego vehicle can reach, and properly assess the risk to take an action.

This paper proposes an algorithm that leverages this intuition from human driving for probabilistic risk assessment and planning by utilizing not only forward but also backward reachability. Since collisions can only happen within the \ac{FRS} of a vehicle, it is unnecessary to predict everything in the occluded region. Instead, staring from a location in the \ac{FRS} of the ego vehicle, one can backward propagate the motion of other vehicles and eventually get all possible initial conditions of risky agents. These initial conditions quantifies and spatially identifies risk-inducing regions and agents, as shown in Figure \ref{fig:demo_init}. With this information, we show how this quantified risk can be used for motion planning for safe navigation in urban environments.

This paper is organized as follows. Section \ref{sec:related_work} reviews related work in reachability-based planning and risk assessment under occlusion. Section \ref{sec:method} shows our problem formulation and proposed risk assessment and motion planning algorithm based on forward  and  backward  reachability. Section \ref{sec:evaluation} describes two baselines and our evaluation metrics. Section~\ref{sec:resutls} describes the quantitative evaluation of the proposed algorithm and illustrates that the proposed method is a significant improvement over existing methods. Section \ref{sec:conclusions} concludes and discusses future directions of this work.
\section{Related Work}
\label{sec:related_work}

Risk assessment is the process of predicting and quantifying the risk of taking a certain action. The proposed method belongs to the category of methods that quantify the risk of having a collision with another agent in the scene as defined in the well-organized survey by \citet{lefevre2014survey}.


Reachability-based motion planning is widely used for autonomous driving~\cite{manzinger2018tactical,orzechowski2018tackling,koschi2017spot,ahn2018reachability,vaskov2019guaranteed,vaskov2019towards,bajcsy2019efficient}. \citet{manzinger2018tactical} assume cooperative agents and proposed an algorithm to negotiate conflicting motions. They try to solve a winner determination problem, which is NP-hard, by restricting the set of combinations to a tree structure. This makes the original problem computationally tractable. \citet{orzechowski2018tackling} over-approximate the \ac{FRS} of incoming traffic, including those who are occluded, by only considering the leading edge of the observable polygon and make sure that the ego vehicle does not collide with the \ac{FRS}. \citet{koschi2017spot} use forward reachability to predict future both the \ac{FRS} of the ego vehicle and the ones of other traffic participants. It performs collision checking of the \acp{FRS} and is evaluated on both urban intersection and highway scenarios. \citet{ahn2018reachability} use both forward and backward reachability, just like the proposed method, to formulate a discrete decision making problem for autonomous driving. \citet{vaskov2019guaranteed} propose \ac{RTD} algorithm for real-time trajectory planning in a static scene by showing how to utilize a \ac{FRS} computed offline during safe online motion planning. \citet{vaskov2019towards} later extended the idea~\cite{vaskov2019guaranteed} to scenes with dynamics agents.

Our approach differs in the following ways. First of all, we formulate the problem probabilistically, whereas the approaches developed by \citet{orzechowski2018tackling,ahn2018reachability,koschi2017spot,vaskov2019guaranteed,vaskov2019towards} do not. The proposed algorithm captures the distribution of risk in both the action space and Cartesian space thus can be more flexible and gives one to potentially adjust the balance between aggressiveness and effectiveness. Secondly, the work by \citet{ahn2018reachability} is intended to be used as a high-level planner for switching between behaviors and thus relies on a low-level planner to generate throttle and steering commands. By contrast, our approach is more like a controller and plans low-level commands directly. Finally, methods by \citet{ahn2018reachability,manzinger2018tactical,vaskov2019guaranteed,vaskov2019towards} do not consider occlusions or other sensory limitations, whereas our approach is designed to operate even under occlusions.

The algorithm proposed in this paper builds upon our previous work~\cite{yu2019occlusion}, which is also a risk assessment and motion planning algorithm. One weakness of our prior work is its poor scalability with respect to the size of the environment. The computational cost is proportional to the area of the occluded regions within the size of the map. The cost of the new approach proposed in this paper does not scale with the area of occluded regions and has a lower collision rate. Also, our previous approach does not identify where to focus the sensory resource.

The main contributions of this paper are as follows:
\begin{itemize}
    \item We propose a probabilistic risk assessment method which identifies the location of the risk-inducing agents and regions.
    \item The proposed method utilizes both forward and backward reachability for efficient risk assessment.
    \item We show how the quantified risk can be used for motion planning for navigating safely in urban environments.
    \item We evaluate our method on $73$ real-world intersections and show quantitative improvement in terms of reduced collision rate when compared to the state of the art.
\end{itemize}

\section{Method}
\label{sec:method}

\begin{figure*}[t]
  \centering
  \subfloat[]{
    \includegraphics[width=0.25\textwidth,trim={30mm 50mm 190mm 40mm},clip]{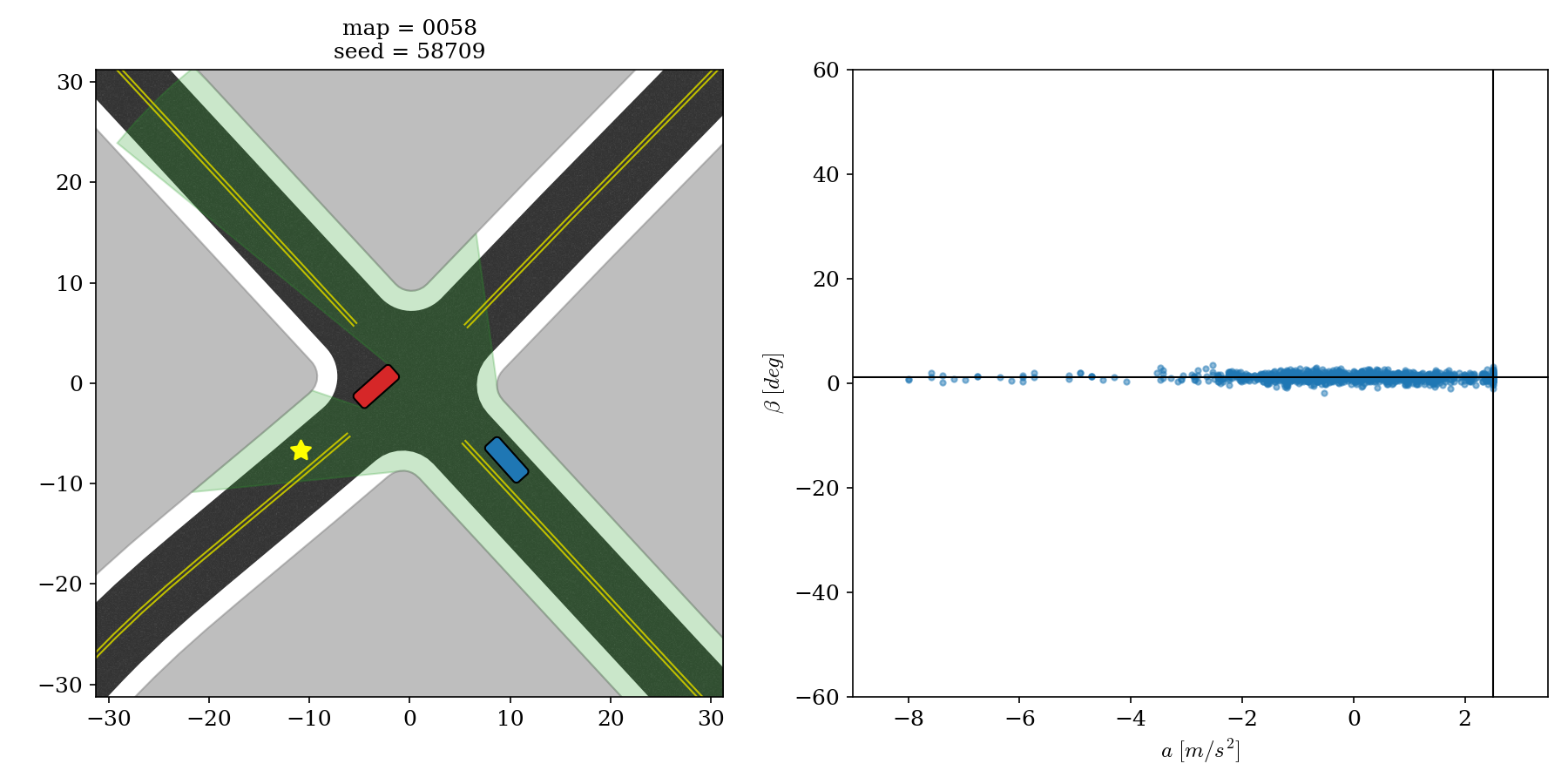}
    \label{fig:step:1}
  }
  \subfloat[]{
    \includegraphics[width=0.25\textwidth,trim={30mm 50mm 190mm 40mm},clip]{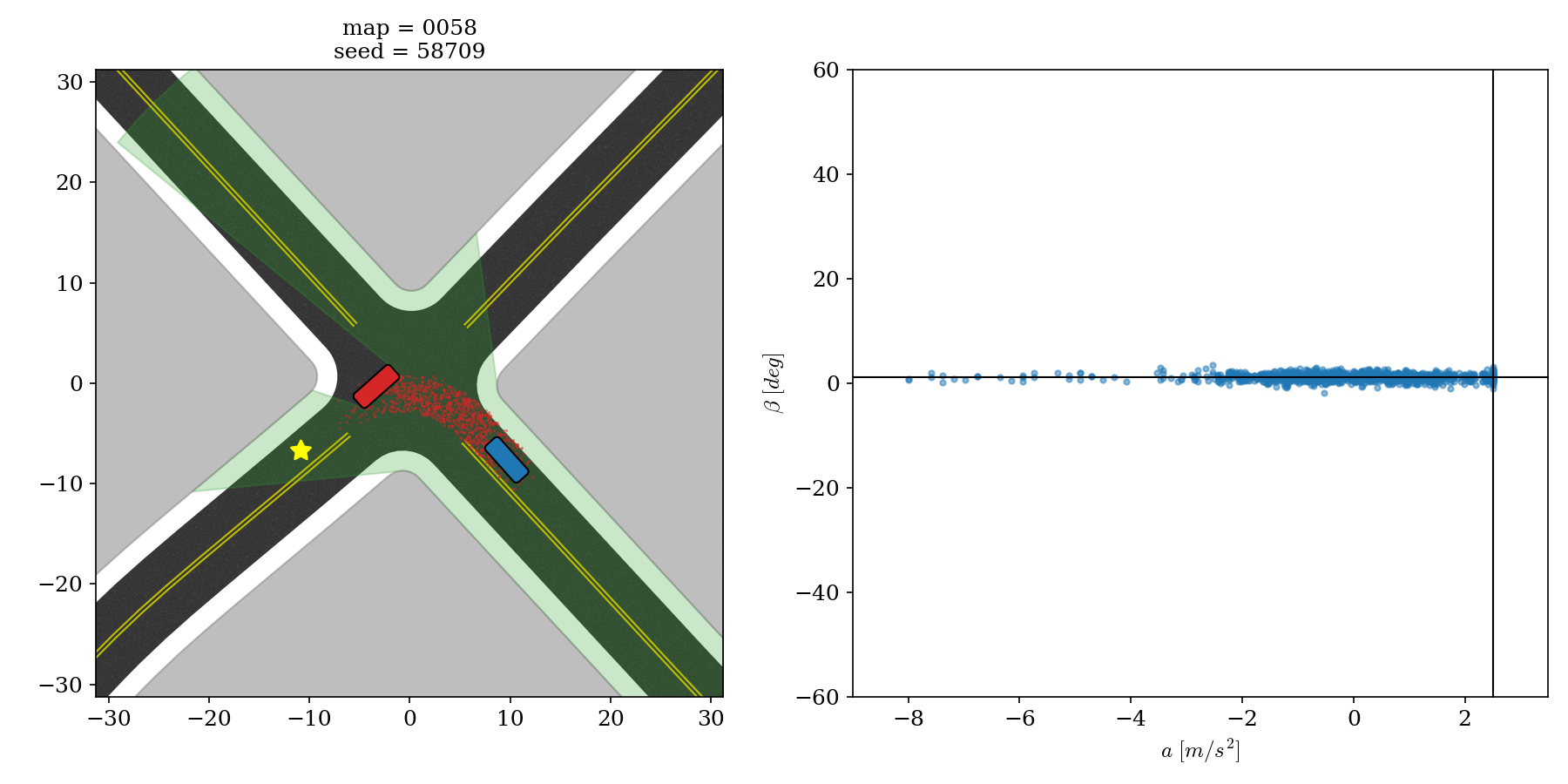}
    \label{fig:step:2}
  }
  \subfloat[]{
    \includegraphics[width=0.25\textwidth,trim={30mm 50mm 190mm 40mm},clip]{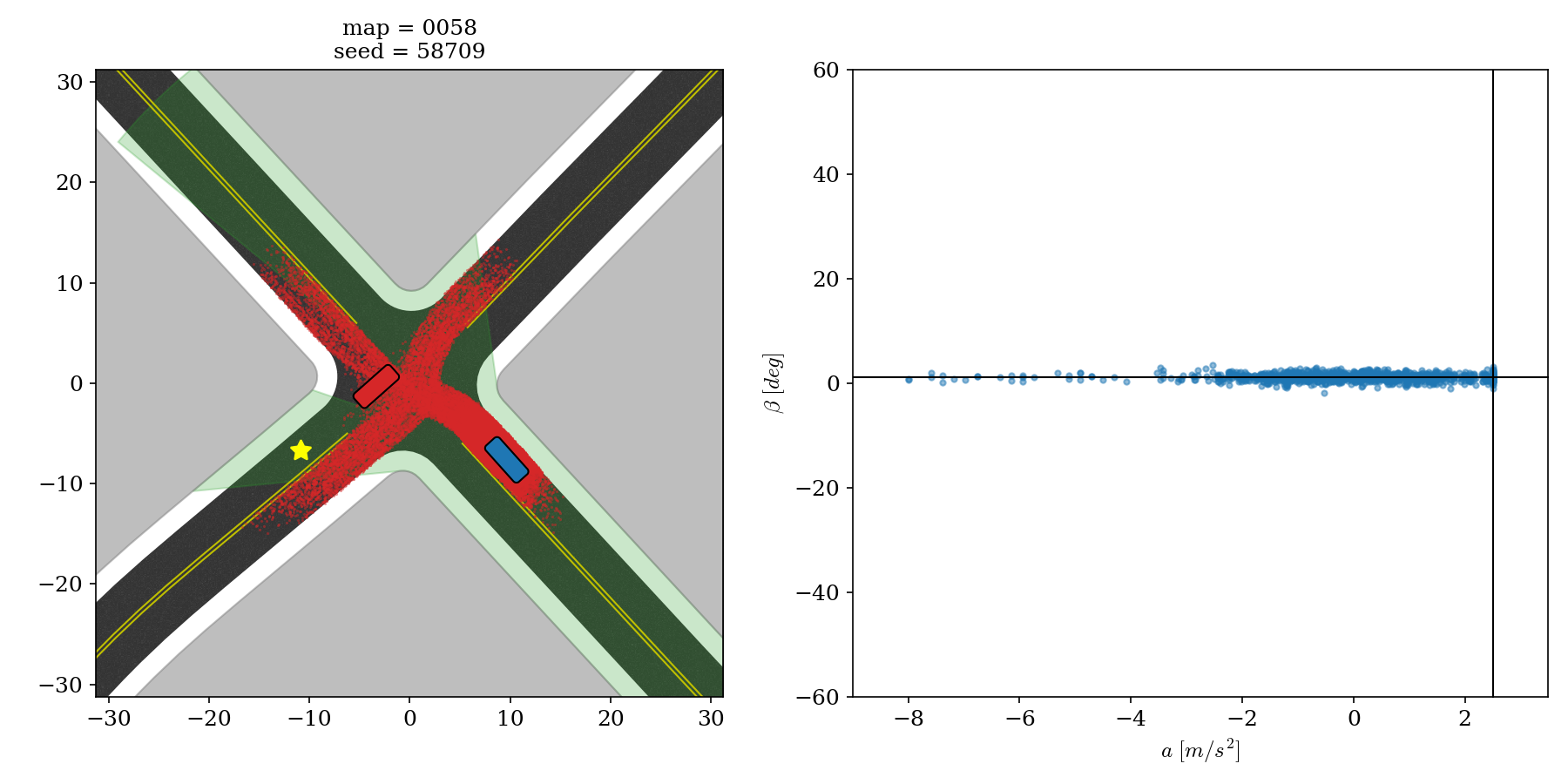}
    \label{fig:step:3}
  }  
  \subfloat[]{
    \includegraphics[width=0.25\textwidth,trim={30mm 50mm 190mm 40mm},clip]{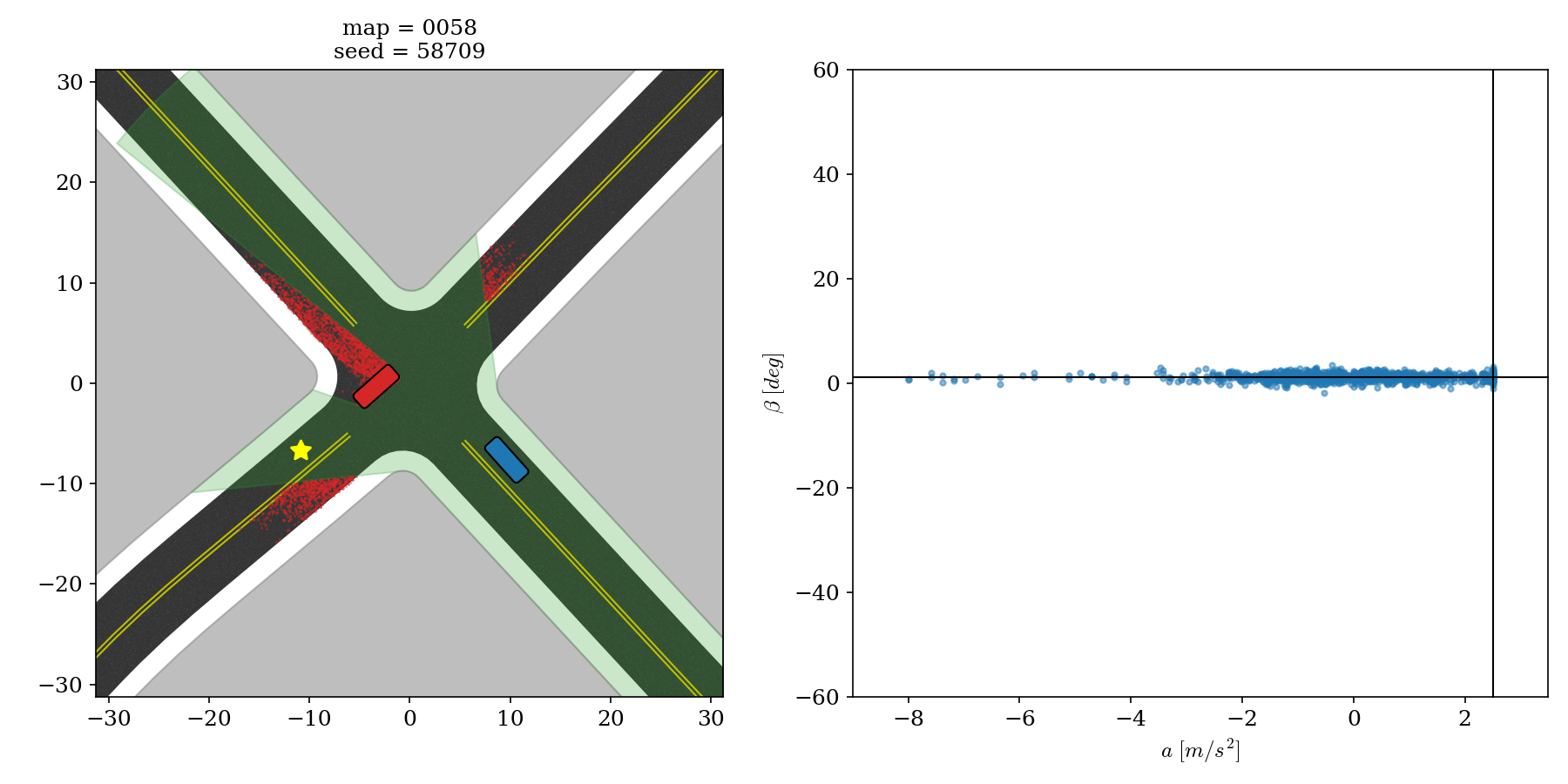}
    \label{fig:step:4}
  }  
  \caption{Step-by-step illustration of different stages of the proposed algorithm. (a) An intersection with occlusion and with one other agent (red box) in the scene. The observable polygon is indicated by the green shaded region, and the goal location is indicated by the yellow star. (b) State 1: calculating FRS$_e$. The dots are projections of the particles $\left\{x_e\left(T^{[i]}\right)\right\}_{i=1}^N$ in 2D. (c) Stage 2: calculating BRS$_o$. The dots are projections of the particles $\left\{x_o^{[i]}(0)\right\}_{i=1}^N$ in 2D. (d) Stage 3: filtering with the observable polygon. The remaining red dots in the figure indicates spatially the source of risk-inducting regions.}
  \label{fig:steps}
\end{figure*}

We first define both forward and backward reachability in Section \ref{sec:reachability}. In Section \ref{sec:risk_assessment}, we show how to use aforementioned reachability for efficient risk assessment. Finally, in Section \ref{sec:planning}, we demonstrate using the quantified risk for motion planning.

\subsection{Bidirectional Reachability}
\label{sec:reachability}
Let $t\in\mathbb{R}$ be time, $\mathcal{X}$ be the state space and $\mathcal{U}$ be the action space. Let $x(t)\in\mathcal{X}$ be the state and $u(t)\in\mathcal{U}$ be the control input of a dynamic system at time $t$. Assume the vehicle's dynamics are described as an \ac{ODE}:
\begin{equation}
    \label{eq:ode}
    \dot{x}(t) = f(t, x(t), u(t)),
\end{equation}
where $f$ is a Lipschitz continuous function.

The \ac{FRS} describes a set of all possible states of a system in the future. Given the dynamics in \eqref{eq:ode}, an initial set $\mathcal{X}_0$ and a fixed time horizon $T$, the \ac{FRS} is defined as
\begin{equation}
\label{eq:frs}
\begin{split}
    \mbox{FRS}(\mathcal{X}_0, T) := \{ &  x(T)\in\mathcal{X} ~|~ x_0 = x(0) \in\mathcal{X}_0, u(t)\in\mathcal{U}, \\
    & \dot{x}(t) = f(t, x(t), u(t)), \forall t\in[0, T] \}
\end{split}
\end{equation}

On the other hand, the \ac{BRS} describes all possible initial states that are able to reach a target set. Given the dynamics in \eqref{eq:ode}, a target set $\mathcal{X}_f$ and a fixed time horizon $T$, the \ac{BRS} is defined as
\begin{equation}
\label{eq:brs}
\begin{split}
    \mbox{BRS}(\mathcal{X}_f, T) := \{ & x(0)\in\mathcal{X} ~|~  x_f = x(T) \in\mathcal{X}_f, u(t)\in\mathcal{U} \\
    & \dot{x}(t) = f(t, x(t), u(t)), \forall t\in[0, T]\}
\end{split}
\end{equation}

Normally, reachable sets are calculated for each observed agent~\cite{vaskov2019guaranteed,vaskov2019towards}. These methods usually perform well when only a limited number of agents are presented in a non-occluded scene. Scenes like urban intersections can still be challenging for these method since predicting \textit{everything}, especially in a occluded environment, can computationally intensive in many cases.
Efforts have been made to ease the calculation under occlusion~\cite{orzechowski2018tackling,yu2019occlusion}, but the complexity still grows along with either the number of occluded regions~\cite{orzechowski2018tackling} or the total area of the occluded regions~\cite{yu2019occlusion}.
For this reason, in Section \ref{sec:risk_assessment} we introduce multiple stages of our algorithm that utilize both forward and backward reachability for efficient risk assessment under occlusion. The complexity of proposed method stays constant no matter how the number or size of the occluded regions grow.

\subsection{Risk Assessment}
\label{sec:risk_assessment}

For autonomous driving, collisions happen only at the intersections of the \acp{FRS} of different agents. In other words, if another agent (occluded or not) cannot reach where the ego vehicle can possibly be in the future, $1.5$ seconds for example, it does not induce risk to the ego vehicle. This observation leads to an important insight -- it is unnecessary to predict the behavior of everything in a scene. Instead, we can focus resources on identifying only those agents who pose a risk to the ego vehicle.

The proposed algorithm, whose behavior is depicted in Figure \ref{fig:steps}, is performed iteratively every $T_r$ seconds, and has a forecast horizon of $T_f$ seconds. It can be broken down into three stages.
\subsubsection{Stage 1 -- \ac{FRS} of the Ego Vehicle}
In the first stage, the \ac{FRS} of the ego vehicle is calculated. In particular, we borrow the idea of using particles from our previous work \cite{yu2019occlusion,jacobs2017real} to represent the \ac{FRS} since it makes the \ac{FRS} probabilistic and the calculation parallelizable.

Without loss of generality, let $t = 0$ at the beginning of each iteration. Let the control inputs be parameterized by some parameters $\theta_e$ and $\theta_o$ such that
\begin{equation}
\label{eq:control}
\begin{split}
    u_e(t) &= u_e\left(t;\theta_e\right)\in\mathcal{U}~~~\forall t \\
    u_o(t) &= u_o\left(t;\theta_o\right)\in\mathcal{U}~~~\forall t,
\end{split}
\end{equation}
where the subscripts $e$ and $o$ indicate quantities tailored to the ego vehicle and other vehicles, respectively. Let the $i-$th particle, $p^{[i]}$, be defined as
\begin{equation}
\label{eq:particle_frs}
    p^{[i]} := \left\{x_e^{[i]}(0), \theta_e^{[i]}(t), \theta_o^{[i]}(t), T^{[i]}\right\}~~~\forall i = 1, \ldots, N,
\end{equation}
where the superscript $[i]$ indicates $i-$th sample of a variable, $x_e(0)$ is ego vehicle's initial state, $T$ is a fixed time horizon. and $N$ is the total number of particles.

In each iteration, we first sample $N$ particles from the following distributions:
\begin{equation}
\label{eq:transition_prob}
\begin{split}
    x_e(0) &\sim U\left(\mathcal{X}_{e,0}\right) \\
    \theta_e &\sim P\left(\theta_e^+~|~\theta_e^-\right) \\
    \theta_o &\sim P\left(\theta_o^+~|~\theta_o^-\right) \\
    T &\sim U\left([0, T_f]\right),
\end{split}
\end{equation}
where $\mathcal{X}_{e,0}$ is the initial set of the ego vehicle, $[\cdot,\cdot]$ is a closed interval, $U(\cdot)$ is an uniform distribution over a set, and $P(\cdot^+~|~\cdot^-)$ is the transition probability of a quantity between subsequent iterations. The transition probabilities enforces smoothness in the action space $\mathcal{U}$ which reduces jittery motion. Since the smoothness is enforced on the distribution level but not directly on the optimal value, actions such as emergency breaking are still allowed. More details about determining the optimal control input from the distribution are described in Section \ref{sec:planning}. 

Now the probabilistic \ac{FRS} of the ego vehicle can be re-written as
\begin{equation}
\label{eq:frs_ego}
\begin{split}
    \mbox{FRS}_e := \Big\{ & x_e\left(T^{[i]}\right)\in\mathcal{X} ~|~ x_e(0) = x_{e,0}^{[i]} \in\mathcal{X}_{e,0} \\
    & \dot{x}_e(t) = f_e\left(t, x_e(t), u_e^{[i]}(t)\right) \\
    & u_e^{[i]}(t)\in\mathcal{U},\forall t\in\left[0, T^{[i]}\right] \Big\}_{i=1}^{N}
\end{split}
\end{equation}

The set FRS$_e$ can be easily calculated by solving $N$ \acp{IVP} in parallel with a regular ODE solver. Note that FRS$_e$ contains $N$ hypotheses of the future states of the ego vehicle, as shown in Figure \ref{fig:step:2}.

\subsubsection{Stage 2 -- \ac{BRS} of Other Agents}
In the second stage of the proposed algorithm, we calculate the \ac{BRS} of other agents in a similar fashion by setting the target set $\mathcal{X}_f$ to FRS$_e$ such that
\begin{equation}
\label{eq:brs_other}
\begin{split}
    \mbox{BRS}_o := \Big\{ & x_o(0)\in\mathcal{X} ~|~ x_e\left(T^{[i]}\right)\in\mbox{FRS}_e \\
    & \mathcal{P}_o\left( x_o\left(T^{[i]}\right) \right) = \mathcal{P}_e\left( x_e\left(T^{[i]}\right) \right) \\
    & \dot{x}_o(t) = f_o\left(t, x_o(t), u_o^{[i]}(t)\right) \\
    & u_o^{[i]}(t)\in\mathcal{U},\forall t\in\left[0, T^{[i]}\right] \Big\}_{i=1}^N,
\end{split}
\end{equation}
where $\mathcal{P}_o:\mathcal{X}_o\rightarrow\mathbb{R}^2$ and $\mathcal{P}_e:\mathcal{X}_e\rightarrow\mathbb{R}^2$ are invertible functions projecting the state spaces into Cartesian space. We use projections since collisions occur when a subset of the states (e.g. positions) overlap. In this case, collision happens at one point in the Cartesian space, while other states such as heading and velocity do not necessary have to be the same and should be assigned separately.

Unlike the FRS$_e$, the BRS$_o$ set cannot be calculated directly with an \ac{ODE} solver. Instead, we need to first reformulate the final value problem into an equivalent \ac{IVP}. For Lipschitz continuous dynamic systems such as \ac{CTRV} model, \ac{CTRA} model, and the bicycle model, the initial value $x_o(0)$ corresponding to particle $i$ can be obtained by solving the following \ac{IVP}:
\begin{equation}
\label{eq:brs_ivp}
\begin{split}
    \dot{z}(t) &= -f_o\left(t, z(t), u_o^{[i]}(t)\right) \\
    z(0) &= x_e\left(T^{[i]}\right)
\end{split}
\end{equation}

The set BRS$_o$ can now be obtained by solving \eqref{eq:brs_ivp} in parallel with a regular ODE solver, and the solution $z\left(t\right)$ is $x_o(t)$ in \eqref{eq:brs_other}, but reversed in time.

BRS$_o$ represents a collection of risky initial conditions of other agents. Each particle corresponds to a possible collision between the ego vehicle and another agent by design, since the target set of BRS$_o$ is chosen to be FRS$_e$. By doing so, this approach can handle occlusion in an efficient way. We no longer need to predict all possible agents in the scene. Instead, the set BRS$_o$ originates only from $\mathcal{X}_{e,0}$ and tells us directly where the potentially dangerous agents are, no matter whether they are occluded or not.
This step is depicted in Figure \ref{fig:step:3}.

\subsubsection{Stage 3 -- Consistency with the Observation}
BRS$_o$ alone does not quantify the risk associated with a particle. We need to incorporate sensory systems, such cameras, radars and \acp{LIDAR}, to identify risky control inputs. A combination of the aforementioned sensors generates a free space around the ego vehicle know as the \textit{observable polygon}, $\mathcal{X}_s$, as shown in Figure \ref{fig:step:4}.

If the entire trajectory of $i-$th particle, $x_o^{[i]}(t)$, ends in $\mathcal{X}_s$, it is considered to be collision-free; otherwise, it is potentially risky. Quantitatively speaking, a weight $w_s^{[i]}$ is assigned to each particle $i$ based on its entire trajectory.
\begin{equation}
\label{eq:w_s}
w_s^{[i]} = \frac{1}{T^{[i]}}\int_0^{T^{[i]}} \mathds{1}\left(x_o^{[i]}(t) \in \mathcal{X}_s \right)dt
\end{equation}
where $\mathds{1}(\cdot)$ is an indicator function. As a results, the weight $w_s^{[i]}$ renders the likelihood of the corresponding action $u_e^{[i]}(t)$ being collision-free or not.

\subsection{Motion Planning}
\label{sec:planning}
Based on the weight $w_s$ from Section \ref{sec:risk_assessment} it is possible to generate a set of safe actions by resampling. Resampling the particles with $w_s$ defined in \eqref{eq:w_s} filters out actions with potential collision.
However, it is insufficient to use just $w_s$ during resampling. For example, in many cases the safest action may be stopping completely before reaching the goal or getting up to the target speed. The ego vehicle can freeze and never reaches the goal. Hence we need another factor to promote forward motion of the ego vehicle.

Let $u_d$ be the desired control input when no risk is presented in the scene. The desired control input can be from a predefined behavior, such as slowing down gently when approaching an intersection, or maintaining the target speed, etc. Assume that $u_d$ is given, we assign another weight $w_d$ to each particle.
\begin{equation}
\label{eq:w_d}
    w_d^{[i]} := \exp\left(-\frac{\delta_u^{[i]2}}{2\sigma_u^2}\right)
\end{equation}
where
$$\delta_u^{[i]2} := \frac{1}{T^{[i]}}\int_0^{T^{[i]}} \left(u_d(t) - u_e^{[i]}(t)\right)^\top\left(u_d(t) - u_e^{[i]}(t)\right) dt$$
and $\sigma_u$ is a hyperparameter for scaling. The weight $w_d$ promotes those control inputs which are closer to the desired one, and thus is able to drive the ego vehicle forward.

A naive way to utilize both $w_s$ for safety and $w_d$ for driving is to interpret them as likelihoods and take their product as the final weight, and resample the particles. However, safe actions might be suppressed when they are no where near the desired action. So taking the direct product may end up with an action that is still too risky.

Instead, we calculate the final weight $w^{[i]}$ in the following way:
\begin{equation}
\label{eq:final_weight}
    w^{[i]} = w_s^{[i]}\cdot\left( \varepsilon\cdot w_d^{[i]} + (1 - \varepsilon)\cdot\left(1 - \min_j w_s^{[j]}\right) \right),
\end{equation}
where $\varepsilon$ is a small number. We set it to $10^{-4}$ in this paper.

To understand \eqref{eq:final_weight} intuitively, let us first consider an extreme case. Assume $w_s^{[i]}=1~\forall i$, indicating that all the trajectories $x_o^{[i]}(t)$ are in the observable polygon $\mathcal{X}_s$ ans thus being collision-free. The weight $w^{[i]}$ becomes proportional to $w_d^{[i]}$, so the actions closer to the desired action $u_d(t)$ will have higher final weights. In this case, resampling with $w^{[i]}$ makes the ego vehicle move according to the desired action.

On the other hand, if \textit{any} of the trajectories shows slight risk (e.g. $1 - \min_j w_s^{[j]} > 10\varepsilon$), the final weight $w^{[i]}$ will be approximately proportional to $w_s^{[i]}$. As a result, safer actions get promoted and the value $w_d^{[i]}$ has negligible effect in this case.
In short, using \eqref{eq:final_weight} with a reasonably small $\varepsilon$ forces the ego vehicle to prioritize safety and to move forward only when it is risk-free.
Resampling with the weight from \eqref{eq:final_weight} gives us a distribution represented by a set of "good" particles. The final step is to select an optimal control input from this distribution for the ego vehicle to execute.

In general, the distribution can be multi-modal, meaning that there may be multiple control inputs that are equally good. For instance, if there are two vehicles approaching an intersection at the same time, two possibly equally good actions for one of the vehicles can be either 1) being aggressive -- accelerating and pass first or 2) being conservative -- waiting until the other one passes. For this reason, we make use of a well-developed clustering algorithm from the machine learning and computer vision community -- \ac{DBSCAN}~\cite{ester1996density} -- to identify multi-modal distributions and select the optimal action.

\ac{DBSCAN} clusters particles in the action space based on the local density without specifying the number of clusters a priori. It identifies the number of cluster automatically, and the centroids of clusters are candidates for the optimal control input. We choose to prioritize safety and make the ego vehicle to be more conservative in this paper. So if there are multiple clusters, we pick the one with the most deceleration. The optimal control input is then executed until the next iteration.

\section{Evaluation}
\label{sec:evaluation}

This section introduces the baselines for the proposed algorithm to be compared against. We also define the metrics for evaluation.

\subsection{Model}
\label{sec:simulation}
Section \ref{sec:risk_assessment} only gives a general description of the pipeline for clarity, but to test and implement the proposed algorithm, we need to pick appropriate models for both the ego vehicle and other vehicles.
Due to its simplicity and tractable dimension, we choose to use the train-like model, meaning that all the agents travels in their own curvilinear coordinates. A concrete example of the model is to have a vehicle traveling on a lane without lane changing. Although we assume no lane changing in this work, it is possible to relax this constraint by using physics-based model like the bicycle model or other behaviour-based models.

By working in curvilinear coordinates, the location of the ego vehicle can be parameterized by only two numbers: longitudinal position $s_e$ along the center line and the lateral offset $b_e$ with respect to the center line. In addition, the heading is completely determined by $s_e$ thus is not necessary to be included as a state. The dynamics of the ego vehicle can be written as
\begin{equation}
\label{eq:ode_ego}
\begin{split}
    \dot{x}_e(t) &= \begin{bmatrix} 0 & 1 \\ 0 & 0 \end{bmatrix} x_e(t) + \begin{bmatrix} 0 \\ 1 \end{bmatrix}u_e(t) \\
    x_e(t) &:= \begin{bmatrix} s_e(t) & \dot{s}_e(t) \end{bmatrix}^\top \\
    u_e(t) &:= a_e~~~\forall t\geq0,
\end{split}
\end{equation}
where $a_e$ is a constant acceleration along the center line. The dynamics of other vehicles is written in a similar way, but assumed to have a constant speed $v_o$ such that
\begin{equation}
\label{eq:ode_other}
\begin{split}
    \dot{x}_o(t) &= u_o(t) \\
    x_o(t) &:= s_o(t) \\
    u_o(t) &:= v_o~~~\forall t\geq0
\end{split}
\end{equation}
If there are multiple valid lanes for non-ego vehicles, each lane should have its own dynamics.

Here we want to emphasize that more sophisticate models can be substituted in directly, but at the cost of higher computational complexity. We choose constant speed/acceleration models because their simplicity and reasonable performance in similar scenarios, as illustrated in our previous work~\cite{yu2019occlusion}. Moreover, since we work in curvilinear coordinates, both models, \eqref{eq:ode_ego} and \eqref{eq:ode_other}, can be approximated by \ac{CTRV} and \ac{CTRA} models, respectively, when the replanning time $T_r$ is small. Both \ac{CTRV} and \ac{CTRA} are shown to have comparable performance in prediction in urban environments~\cite{schubert2008comparison}.

The desired action $u_d(t)$ is designed to have a constant deceleration when approaching the stopline, and to switch to a saturating proportional controller to track a target speed.

\subsection{Baseline 1}
\label{sec:baseline1}
The first baseline is an occlusion-aware algorithm by \citet{orzechowski2018tackling}. We recreate and simulate two of the scenarios demonstrated in their paper. The map is a four-way unsignalized intersection partially occluded by a non-transparent static object.

The first scenario consists of no other agent, and the second one has one other agent coming from the left hand side of the ego vehicle. In both scenarios, the goal is to navigate safely through the intersection and to reach a target speed of $9~m/s$. The geometries of the map are measured using AutoCAD, and the velocity of the other agent is estimated afterwards.

\subsection{Baseline 2}
\label{sec:baseline2}
The second baseline is another occlusion-aware method from our previous work \cite{yu2019occlusion}. We simulate the same set of $1000$ random scenarios with up to five other agents for each scene on both the proposed method and the second baseline. Each scene is an four-way unsignalized intersection and the geometries are extracted from \ac{OSM}. There are $73$ real-world intersections in total. In addition, there is also a synthetic layout where the roads are straight and perpendicular to each other.

These maps have significantly more severe occlusion comparing to those in the first baseline \cite{orzechowski2018tackling}. Buildings are added to the map with a two-meter buffer around the driving surface, as shown in Figure \ref{fig:demo_init} and \ref{fig:steps}. Speed of a non-ego agent is randomly set to a constant between $4$ to $12~m/s$. For simplicity, we assume no collision between non-ego agents and they do not interact with other agents.

\begin{figure*}[t]
    \centering
    \subfloat[]{
        \includegraphics[width=0.18\textwidth,trim={0mm 0mm 160mm 12mm},clip]{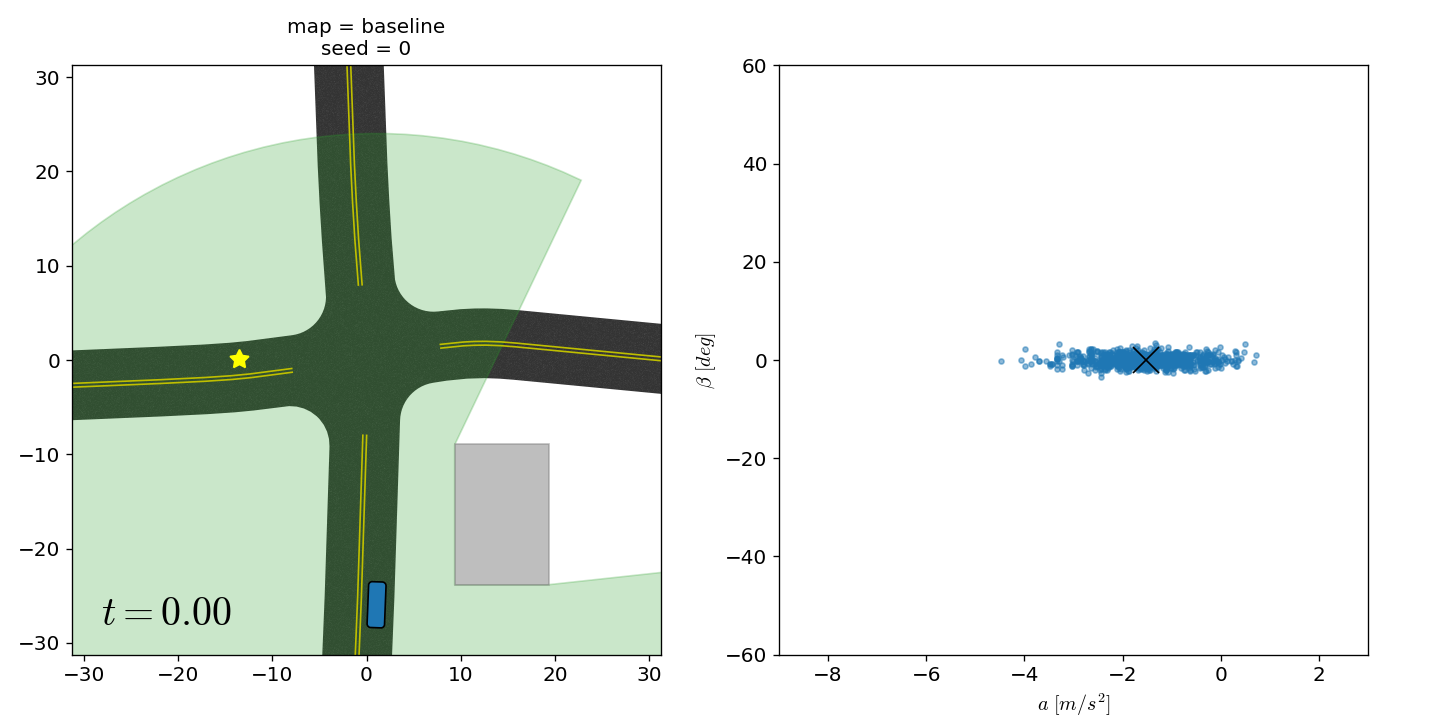}
        \label{fig:scene1:t0}
    }
    \subfloat[]{
        \includegraphics[width=0.18\textwidth,trim={0mm 0mm 160mm 12mm},clip]{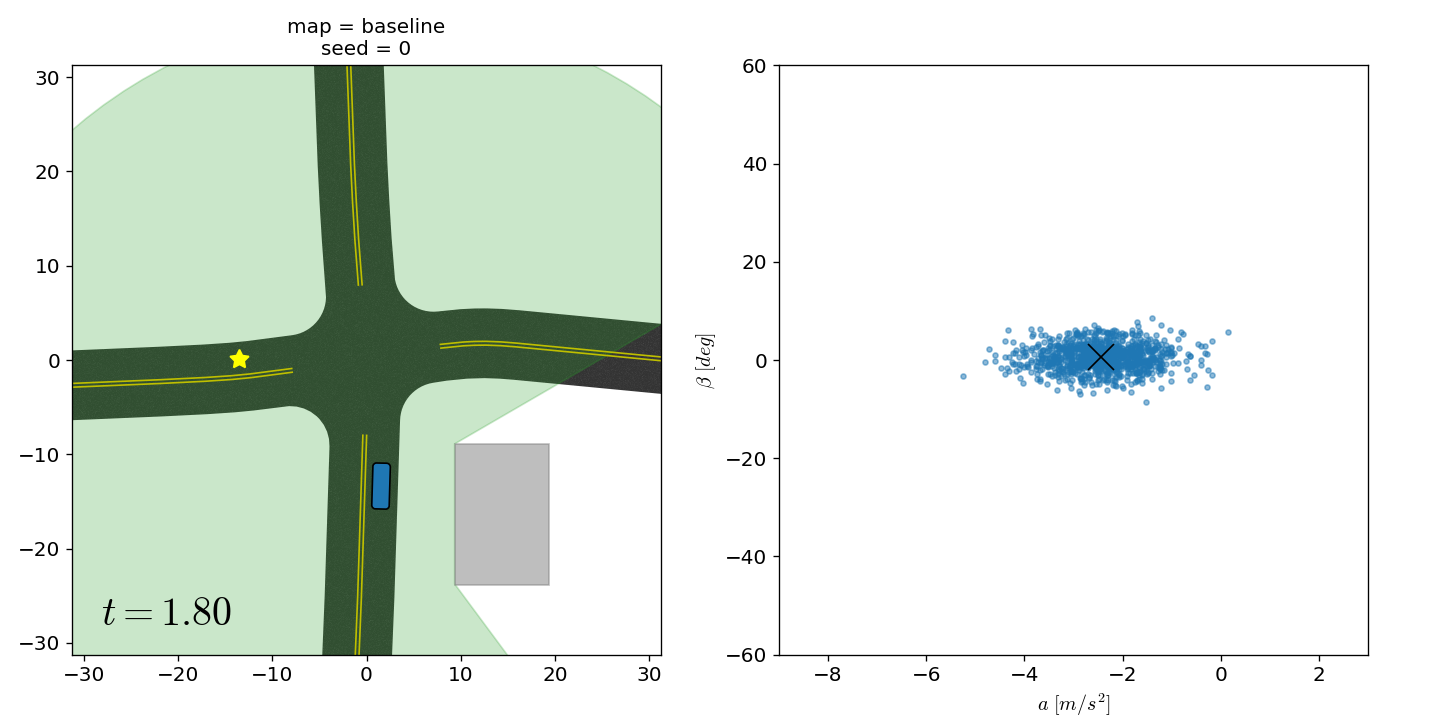}
        \label{fig:scene1:t1}
    }
    \subfloat[]{
        \includegraphics[width=0.18\textwidth,trim={0mm 0mm 160mm 12mm},clip]{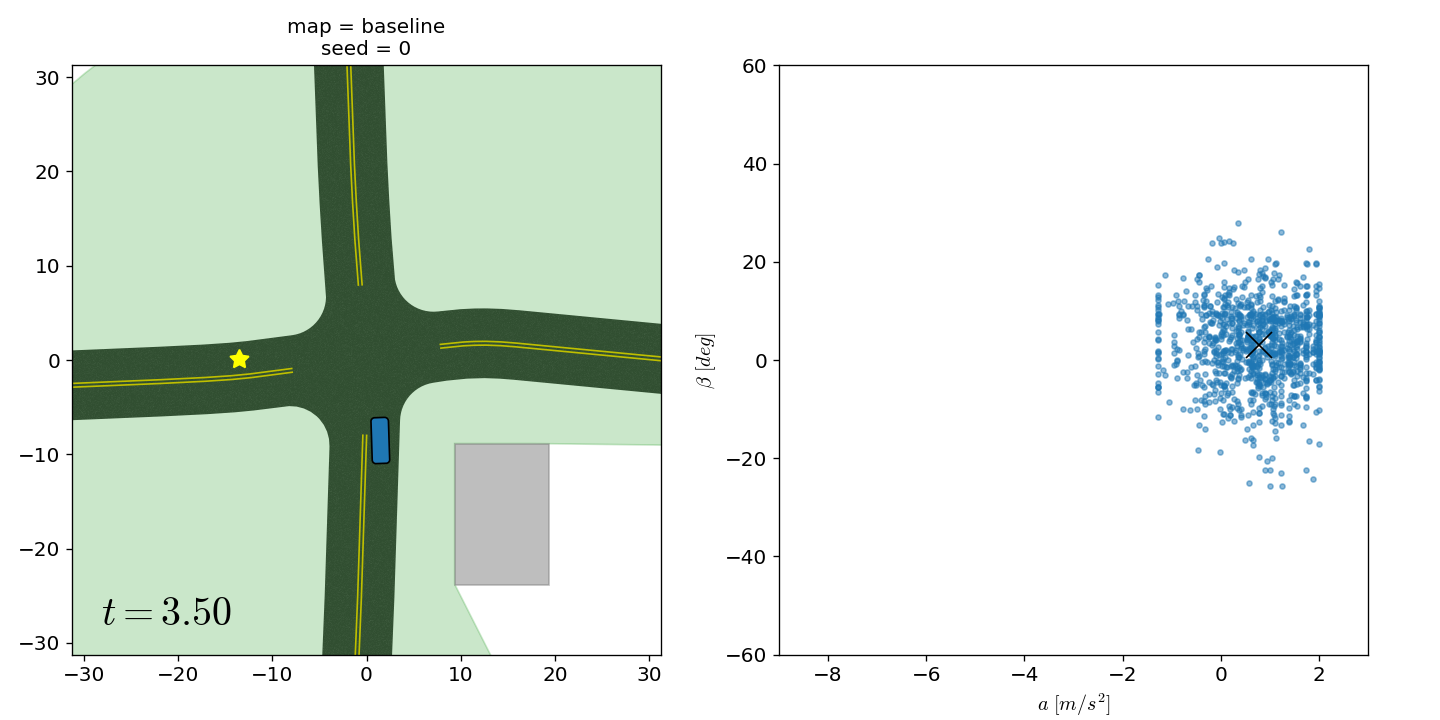}
        \label{fig:scene1:t2}
    }
    \subfloat[Baseline 1~\cite{orzechowski2018tackling}]{
        \includegraphics[width=0.22\textwidth,trim={0mm 0mm 0mm 0mm},clip]{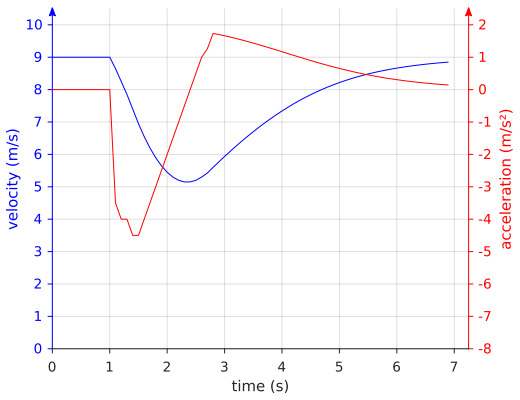}
        \label{fig:scene1:baseline:profile}
    }
    \subfloat[Proposed]{
        \includegraphics[width=0.24\textwidth,trim={0mm 3mm 0mm 0mm},clip]{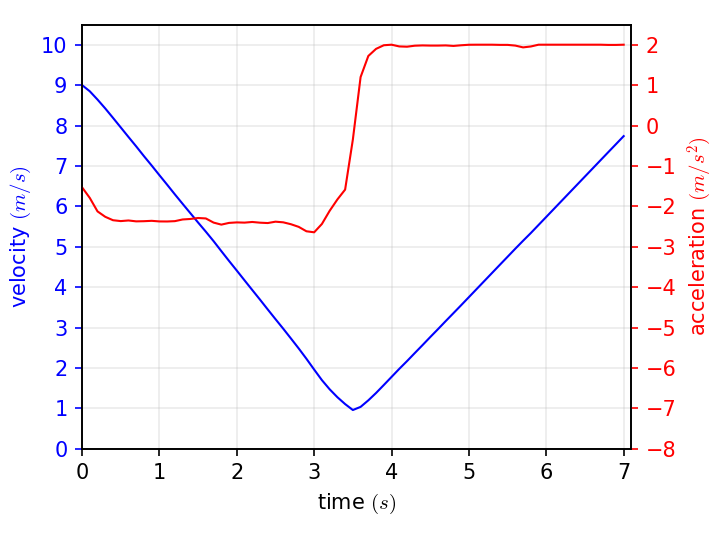}
        \label{fig:scene1:proposed:profile}
    } \\
    \subfloat[]{
        \includegraphics[width=0.18\textwidth,trim={0mm 0mm 160mm 12mm},clip]{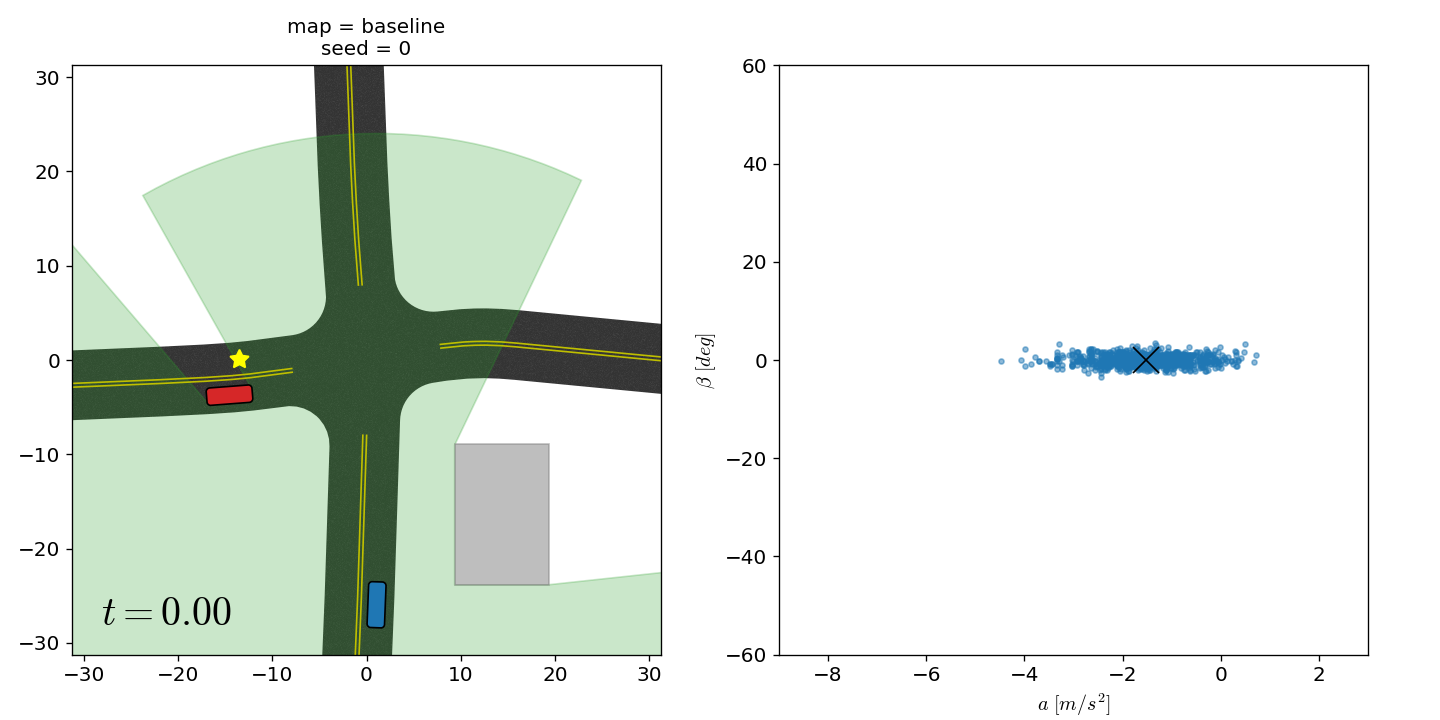}
        \label{fig:scene2:t0}
    }
    \subfloat[]{
        \includegraphics[width=0.18\textwidth,trim={0mm 0mm 160mm 12mm},clip]{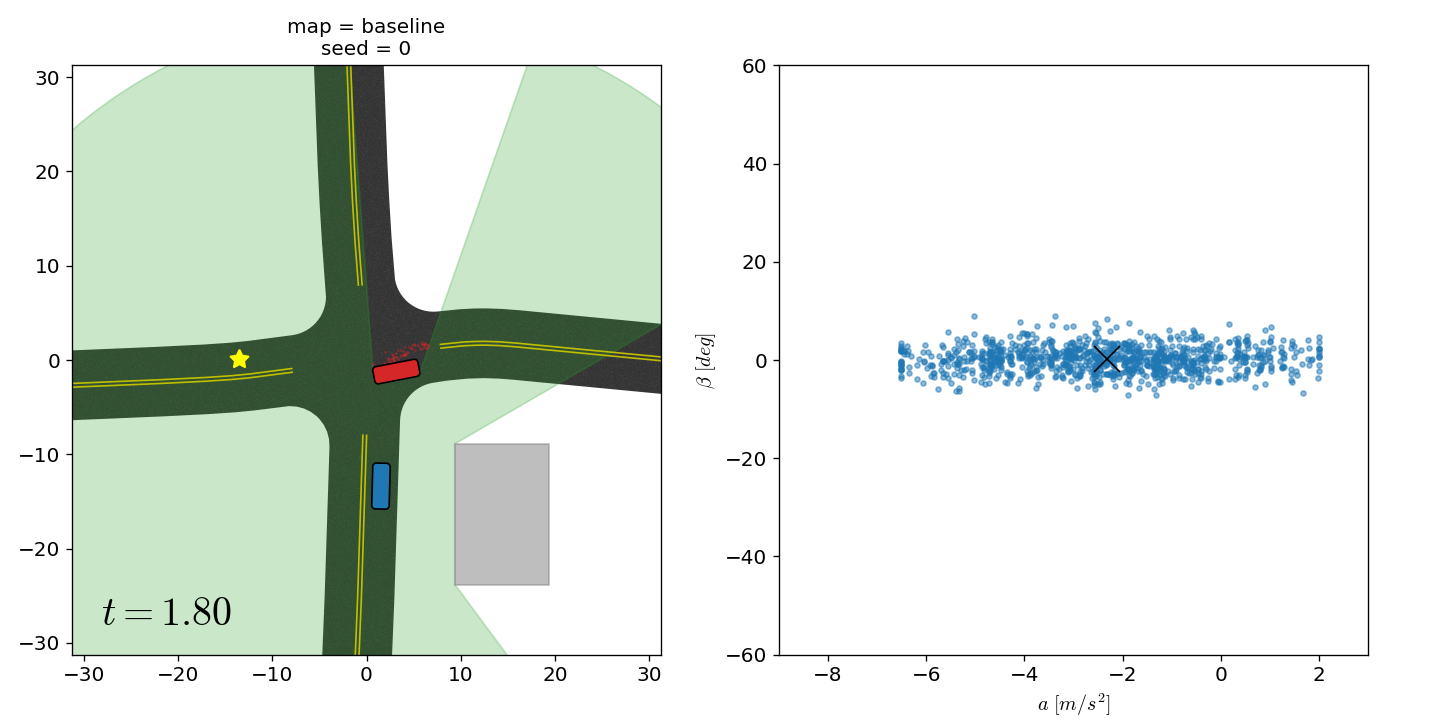}
        \label{fig:scene2:t1}
    }
    \subfloat[]{
        \includegraphics[width=0.18\textwidth,trim={0mm 0mm 160mm 12mm},clip]{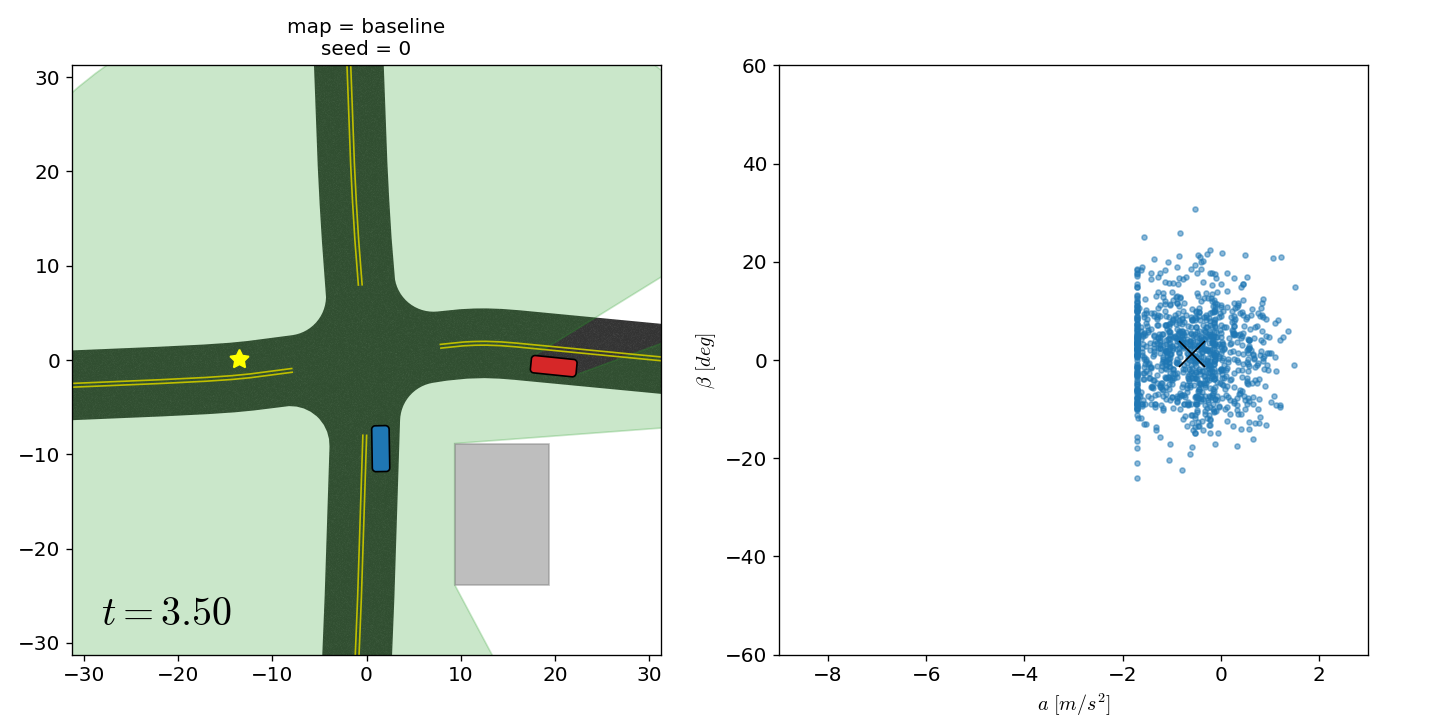}
        \label{fig:scene2:t2}
    }
    \subfloat[Baseline 1~\cite{orzechowski2018tackling}]{
        \includegraphics[width=0.22\textwidth,trim={0mm 0mm 0mm 0mm},clip]{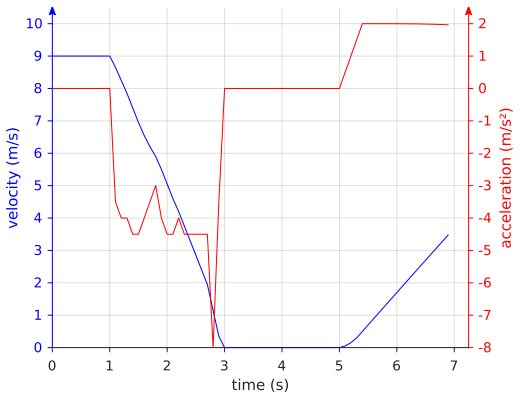}
        \label{fig:scene2:baseline:profile}
    }
    \subfloat[Proposed]{
        \includegraphics[width=0.24\textwidth,trim={0mm 3mm 0mm 0mm},clip]{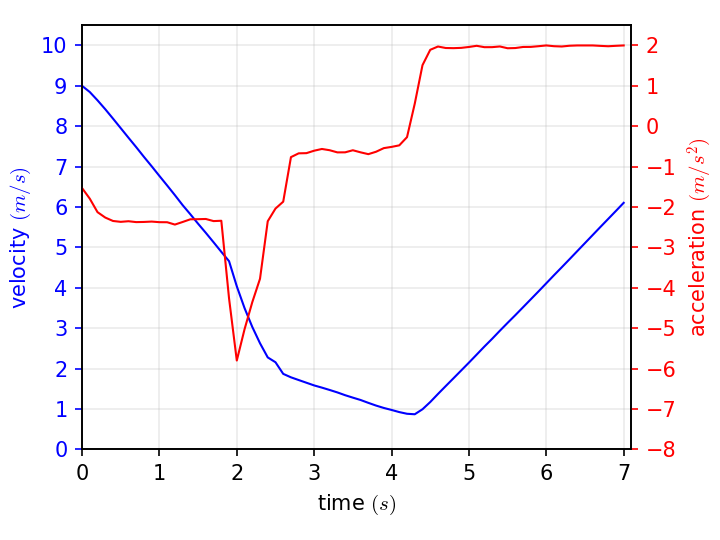}
        \label{fig:scene2:proposed:profile}
    }
    \caption{
        Two scenarios in the first baseline~\cite{orzechowski2018tackling}. The first scenario (top row) is in a map with one static obstacle (gray) and no incoming traffic, and the second scenario (bottom row) is in the same map with one other vehicle (red) coming from the left. The first three columns (a-c, f-h) show key frames of the proposed method, and the other two columns (d,e,i,j) show the speed and acceleration profiles. In the first scenario, the baseline abruptly breaks at $t=1.0s$ whereas the proposed method maintains a near constant deceleration, as shown in (d) and (e). In the second scenario, the baseline breaks at both $t=1.0s$ and $t=2.8s$, whereas the proposed method only breaks once at $t=1.9s$, as shown in (i) and (j). In addition our method does not stop completely in the second scenario. Instead, it creeps forward between $t=2.7s$ and $t=4.3s$ to gather more information. This behavior results in a higher final speed and thus makes the ego vehicle faster to reach the goal. The fourth column is adopted directly from Figure 4 in~\cite{orzechowski2018tackling}.
    }
    \label{fig:bl1}
\end{figure*}

\subsection{Metrics}
\label{sec:metrics}
For the first baseline, we investigate the speed and acceleration profiles for both ride comfort and efficiency. Ride comfort is evaluated by the number of abrupt breaking and the maximum deceleration, whereas efficiency is indicated by the terminal speed of the ego vehicle.

Collision rate is evaluated for the second baseline on one synthetic and all the $73$ real-world intersections. This results in $74000$ simulations in total for each method.
\section{Results}
\label{sec:resutls}


\subsection{Baseline 1}
Figure \ref{fig:bl1} shows the behaviors under two scenarios recreated from the first baseline~\cite{yu2019occlusion}. Both the first baseline and the proposed method enable the ego vehicle to reach to the goal without collision in either scenario. The ego vehicle slows down when approaching the intersection, no matter if there is any incoming traffic.


However, they have three key differences. To begin with, in the first scenario the baseline~\cite{orzechowski2018tackling} has a sudden break at $t=1.0s$, while the proposed method stays roughly at a constant deceleration. In addition, the maximum breaking (i.e. deceleration) in the second scenario is $27.5\%$ lower in the proposed method. The reduced breaking and jerk indicate higher ride comfort. Finally, in the second scenario the ego vehicle creeps forward between $t=2.7s$ and $t=4.3s$ in order to get a clearer view, as shown in Figure \ref{fig:scene2:t2} and \ref{fig:scene2:proposed:profile}. Once it gets enough information and thinks it is risk-free, it then proceeds. This behavior leads to a higher efficiency, meaning that it obtains a $57.9\%$ higher speed at the end of the simulation all without sacrificing safety. 

\subsection{Baseline 2}

\begin{table}
    \centering
    \begin{tabular}{@{}rrr@{}} \toprule
        & median collision rate & \# of zero-collision intersections \\ \midrule
        Baseline 2~\cite{yu2019occlusion} & $1.70\%$ & $0$ out of $74$ \\
        Proposed & $0.20\%$ & $10$ out of $74$ \\ \bottomrule
    \end{tabular}
    \caption{Simulation results at $74$ intersections.}
    \label{tbl:bl2}
\end{table}

Among the $74$ intersections, the proposed method obtains a median collision rate of $0.20\%$, which is a significant reduction compared to the $1.70\%$ of the second baseline~\cite{yu2019occlusion}, as shown in Table \ref{tbl:bl2}. Moreover, the proposed method achieves zero collision at not only the synthetic layout, but also $9$ real-world layouts. In contrast, the baseline~\cite{yu2019occlusion} has a minimum collision rate of $0.70\%$ at one of the $74$ intersections.
\section{Conclusions and Future Work}
\label{sec:conclusions}

We propose a probabilistic risk assessment method for motion planning under occlusion in urban intersections. The proposed algorithm is able to quantify the distribution of risk in the action space, and can be used to generate low-level control inputs for an autonomous vehicle to navigate safely. It also identifies spatially where the risk-inducing regions are via both forward and backward reachability. The proposed method is compared quantitatively to previous work and shows significant improvement in terms of both efficiency and safety.

Future work include using the proposed method to redirect and focus sensory resources to further reduce impact on limited sensing, formulating and solving the problem analytically without sampling to obtain safety guarantees, and deploying it to hardware platforms for testing in the real-world.

\newpage
\printbibliography

\end{document}